\documentclass[final,3p,times]{elsarticle}
\usepackage{multicol}
\usepackage{multirow}
\usepackage{amsmath}
\usepackage{url}
\usepackage{graphics}
\usepackage{float}
\usepackage{caption}
\usepackage{subcaption}
\usepackage{booktabs}
\usepackage{listings}
\usepackage{amssymb}

\begin{document}
	
\begin{frontmatter}
		
		

\title{Long Is More Important Than Difficult for Training Reasoning Models}

\author[label1]{Si Shen}
\author[label1]{Fei Huang}
\author[label2]{Zhixiao Zhao}
\author[label2]{Chang Liu}
\author[label2]{Tiansheng Zheng}
\author[label3]{Danhao Zhu\corref{cor1}}
\cortext[cor1]{Corresponding author}
\address[label1]{Nanjing University of Science and Technology, Nanjing, Jiangsu, 210094, China}
\address[label2]{Nanjing Agricultural University, Nanjing, Jiangsu, 210095, China}
\address[label3]{Criminal Science and Technology, Jiangsu Police Institute, Nanjing, Jiangsu, 210031, China}

\begin{abstract}
Difficult problems, which often result in long reasoning traces, are widely recognized as key factors for enhancing the performance of reasoning models. However, such high-challenge problems are scarce, limiting the size of available datasets. In this paper, we propose a simple method to decouple the reliance on problem difficulty. First, we empirically demonstrate that reasoning length, rather than problem difficulty, primarily influences the performance of trained models. Second, we identify a scaling law on reasoning length, showing that model performance increases in a log-linear fashion as the reasoning data length grows. Finally, we introduce a straightforward technique to generate reasoning data of arbitrary length, and show that synthesized data is effective for training reasoning models. After fine-tuning the Qwen2.5-32B-Instruct language model on our Long1K dataset, we present our model, Long1K-32B, which achieves remarkable performance with only 1,000 training samples, achieving 95.6\% accuracy on MATH, and
71.1\% on GPQA outperforming DeepSeek-R1-Distill-Qwen-32B. The model, code, and dataset are all open-sourced, available at https://huggingface.co/ZTss/LONG1.
\end{abstract}


\end{frontmatter}

\section{Introduction}
\label{intro}

Large Language Model (LLM)-based reasoning models have demonstrated impressive capabilities across various benchmarks, often producing extended sequences of reasoning before arriving at a final answer. Notable models include OpenAI o1~\cite{jaech2024openai}, DeepSeek-R1~\cite{guo2025deepseek}, and Gemini Flash Thinking~\cite{deepmind2024gemini}. Among them,  DeepSeek-R1~\cite{guo2025deepseek} offers publicly accessible solutions, demonstrating how reinforcement techniques can be applied to a large instruction-based model. They also show that reasoning data produced by a strong teacher model can be used to effectively train smaller models with robust reasoning abilities.

Training a reasoning model—whether through reinforcement learning or by distilling knowledge from a large model, has relied on highly challenging problems. For example, STILL-2~\cite{Min2024ImitateEA} reports that extremely difficult tasks are critical for self-improvement in reasoning models, while DRT-o1~\cite{Wang2024DRTDR} employs tough translation problems that yield complex, extended reasoning traces. In the context of knowledge distillation, both s1~\cite{Muennighoff2025s1ST} and LIMO\cite{Ye2025LIMOLI} emphasize the importance of challenging problems; they demonstrate that, with fewer than 1,000 difficult examples, a smaller model can even surpass the o1-preview~\cite{openai2024reasoning} model.

However, in real-world scenarios, truly difficult problems are scarce. For instance, existing datasets frequently contain only a small fraction of samples (e.g., 2.35\% in the Sky-T1-32B-Preview~\cite{Li2025STT} training dataset, and 7.34\% in the LIMO~\cite{Ye2025LIMOLI} training dataset) with reasoning lengths exceeding 15k tokens. This scarcity inherently restricts the size of training sets and, consequently, the potential for improving reasoning models at scale.

The prevailing assumption behind the efficacy of difficult tasks is that they induce more complex reasoning chains, enabling LLMs to leverage their pre-trained knowledge more effectively~\cite{Ye2025LIMOLI}. Yet, this assumption remains largely untested. While difficult problems do tend to produce more extended reasoning, we hypothesize that reasoning length itself, rather than the problems’ difficulty, is the key driver of improved model performance. In Section 2, we present two sets of experiments that support this view: when the total reasoning length is similar, a dataset composed of easier yet more exhaustive questions achieves comparable performance to one featuring conventionally “hard” problems.

To further explore the role of reasoning length, we investigate its effect in Section 3, where we find that longer training data consistently leads to better performance. Inspired by this, we propose a straightforward approach to generate arbitrarily long reasoning sequences. We simply extend the reasoning length without the need for complex and labor-intensive data filtering. We have released the synthesized dataset Long1K, which comprises a total of 1000 samples, with 800 samples having a reasoning length of 32k tokens, and the remaining 200 samples having a random reasoning length. After fine-tuning Qwen2.5-32B-Instruct~\cite{Yang2024Qwen25TR} on Long1K, our Long1K-32B model achieves an accuracy of 95.6\% on the MATH500~\cite{Hendrycks2021MeasuringMP} test and achieves 71.1\% accuracy on GPQA Diamond~\cite{Rein2023GPQAAG} using only 1,000 training samples, matching the performance level of DeepSeek-R1 and surpassing other baseline methods by a notable margin.
Our primary contributions can be summarized as follows:

\begin{itemize}
	\item \textbf{Challenging a common assumption:} We question the prevalent belief that problem difficulty is the most critical factor. Instead, our experiments suggest that \emph{reasoning length} is key to training high-performance reasoning models. This insight allows us to build large-scale, long-reasoning datasets without being constrained by the rarity of extremely difficult problems.
	
	\item \textbf{Identifying a scaling law on reasoning length:} We observe that model performance improves nearly linearly as the length of training data increases exponentially. This phenomenon highlights the efficiency gains achievable by focusing on the length of reasoning sequences.
	
	\item \textbf{Proposing a simple synthesis method:} We introduce a technique to generate arbitrarily long reasoning data. Using this method, we release the \emph{Long1K} dataset, upon which our \emph{Long1K-32B} model is fine-tuned. This model surpasses existing baselines on benchmarks such as MATH500 and GPQA Diamond, demonstrating that extended reasoning sequences can significantly enhance model performance.
\end{itemize}

\section{Reasoning Length is More Important than Difficulty}
\subsection{Hypotheses}
The conventional wisdom assumes that difficult problems are essential for training reasoning models~\cite{Lightman2023LetsVS}. This assumption stems from the observation that difficult problems often require long reasoning traces. However, we hypothesize that reasoning length itself, rather than problem difficulty, is the key factor driving model performance. 

To validate this hypothesis, we design experiments where datasets contain similar reasoning lengths but differ in difficulty. If problem difficulty were the dominant factor, models trained on easier datasets should perform worse. Instead, we find that models trained on long but easier problems perform comparably to those trained on difficult problems, challenging the necessity of problem difficulty in reasoning dataset construction.

Our core experiment examines the effects of reasoning length and problem difficulty on model performance by comparing two distinct datasets. To this end, we constructed the following two types of datasets:
\begin{itemize}
	\item \textbf{Long Problems:}  Each problem consists of several sub-questions, but each sub-question is simple. Due to the extensive content, the reasoning length is long.
	\item \textbf{Difficult Problems:} There is only one problem, but due to its complexity, the reasoning length is long.
\end{itemize}

To maintain experimental fairness, we carefully balanced the two datasets. They have similar reasoning lengths and contain the same number of training samples, but differ significantly in problem difficulty. We constructed these two types of datasets using distinct methodologies and conducted the following two main experiments. All experiments were carried out using the Qwen2.5-7B-Instruct and Qwen2.5-32B-Instruct models, with evaluations conducted on the AIME2024~\cite{aime} and MATH500~\cite{Hendrycks2021MeasuringMP} datasets.

\begin{table*}[!t]
	\centering
	\setlength{\tabcolsep}{6pt}
	{
		\fontsize{10}{13}\selectfont 
		\renewcommand\arraystretch{1.5}
		\resizebox{\linewidth}{!}{
			\begin{tabular}[l]{p{0.5\linewidth}p{0.8\linewidth}} 
				\toprule
				\textbf{Problem} & \textbf{Example} \\ \midrule
				\textbf{Long Problems(Composite)} & \small Consider a convex pentagon with side lengths \(x, y, z, w, v\), where \(xyzwv = 1\). The pentagon can be divided into three triangles by drawing perpendiculars from one vertex. The areas of these triangles are proportional to the side lengths adjacent to the vertex. 
				\textbf{1.} Given that \(x^2 \leq 4\), find the maximum possible area of the pentagon. 
				\textbf{2.} How many distinct sets of positive integer side lengths satisfy the given conditions? \\ \midrule
				\textbf{Difficult Problems(Composite)} & \small Determine the range of possible values for the angle \( \theta \) such that the rotation matrix \( R(\theta) = \begin{pmatrix} \cos \theta & -\sin \theta \\ \sin \theta & \cos \theta \end{pmatrix} \) satisfies the condition that the quadratic equation has no real roots, given the dimensions of a rectangular box whose length, width, and height are the arithmetic mean, geometric mean, and harmonic mean of two distinct positive real numbers \( x \) and \( y \), respectively. \\ \midrule
				\textbf{Long Problems(Double)} & \small \textbf{Question 1} is "48 blacksmiths need to shoe 60 horses. What is the minimum time they will spend on the job if each blacksmith takes 5 minutes per horseshoe?". \textbf{Question 2} is "If 25,197,624 hot dogs are packaged in sets of 4, how many will be left over?" \\ \midrule
				\textbf{Difficult Problems(Single)} & \small \textbf{Question 1} is "A straight ladder \(A B\) of mass \(m=1 \mathrm{~kg}\) is positioned almost vertically such that point \(B\) is in contact with the ground with a coefficient of friction \(\mu=0.15\). It is given an infinitesimal kick at the point \(A\) so that the ladder begins rotating about point \(B\). Find the value \(\phi_{m}\) of angle \(\phi\) of the ladder with the vertical at which the lower end \(B\) starts slipping on the ground." \\ \bottomrule
				\end{tabular}
			}
			\caption{Examples of Four Different Problem Types: Long Problems(Composite), Difficult Problems(Composite), Long Problems(Double), and Difficult Problems(Single).}
			\label{tab:1}
		}
	\end{table*}

\subsubsection{Long Problems(Composite) vs. Difficult Problems(Composite)}

To ensure that long problems and difficult problems are based on a similar distribution of concepts and creation process, and that the synthesized questions themselves conform to common types of long problems and difficult problems, we adopted a specific approach. We extracted mathematical knowledge points from multiple mathematics collections and synthesized them to form two datasets, with the detailed extraction and synthesis methods provided in the Appendix A.

\begin{itemize}
	\item \textbf{Long Problems(Composite):} A comprehensive problem constructed using mathematical knowledge points, with each problem containing several sub-problems, each of which is relatively easy.
	\item \textbf{Difficult Problems(Composite):} A single complex mathematical problem integrates multiple mathematical  knowledge points.
\end{itemize}

\begin{figure}[h]
	\centering
	\includegraphics[width=1\textwidth]{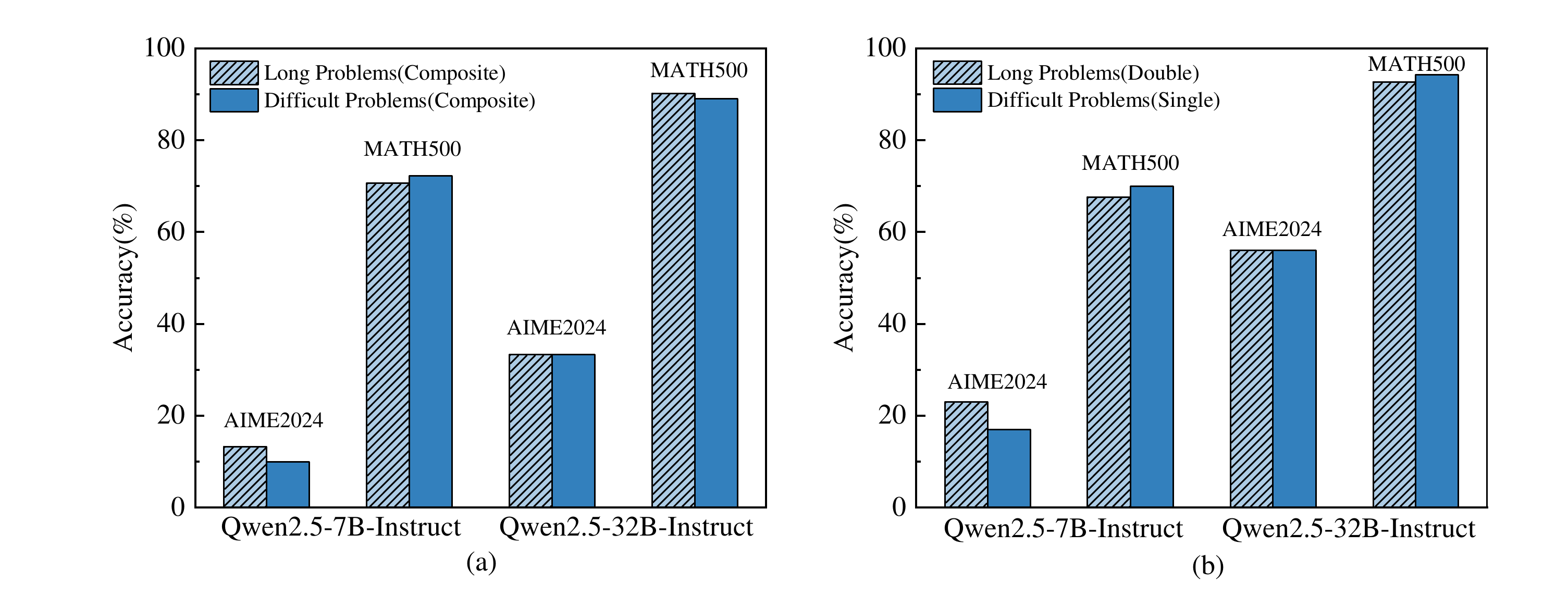}
	\caption{(a) Accuracy comparison of Qwen2.5-7B-Instruct and Qwen2.5-32B-Instruct after training on Long Problems(Composite) and Difficult Problems(Composite). (b) Accuracy comparison of Qwen2.5-7B-Instruct and Qwen2.5-32B-Instruct after training on Long Problems(Double) and Difficult Problems(Single).}
	\label{fig:2.1}
\end{figure}

If problem difficulty is essential, we would expect models trained on difficult problems to outperform those trained on easier, multi-step problems. However, as shown in Figure~\ref{fig:2.1}(a), the results shows that their performance is nearly identical. Notably, in the AIME2024 test, the 7B model trained on simpler multi-step problems achieved an accuracy of 13.3\%, outperforming the 10\% accuracy of the model trained on difficult problems. Despite lacking inherently difficult problems, models trained on long multi-step questions perform equally well. This challenges the common belief that problem difficulty is the main factor influencing model performance. Instead, it suggests that reasoning length plays a more decisive role, underscoring its importance in improving model capabilities.

\subsubsection{Long Problems(Double) vs. Difficult Problems(Single)}
To further test whether reasoning length is more influential than problem difficulty, we expanded our experiment by constructing two datasets in a more direct manner and compared the following two types of problems:
\begin{itemize}
	\item \textbf{Long Problems(Double):} Each query contains two independent problems, effectively doubling the reasoning length.
	\item \textbf{Difficult Problems(Single):} A single complex problem per query.
\end{itemize}
	
As shown in Figure~\ref{fig:2.1}(b), models trained on Long Problems(Double) and Difficult Problems(Single) perform similarly. On the AIME 2024 test, both the 32B model trained on Long Problems(Double) and trained on Difficult Problems(Single) achieved an accuracy of 33.3\%. Even for the 7B model, the performance of the model trained on Double Problems was superior. The results again confirm that increasing reasoning length—even with independent problems—consistently improves performance. It further supports the idea that performance gains arise primarily from increased reasoning length rather than problem difficulty.

\subsection{Summary}

From these experiments, we conclude that reasoning length plays a more crucial role than problem difficulty. These findings suggest that rather than focusing on difficult problems, one can achieve similar or superior results by simply increasing reasoning length through structured problem decomposition or concatenation, without necessarily increasing problem difficulty. In particular, longer reasoning chains significantly improve model performance on complex tasks, even when the problems themselves are not inherently challenging.

Since reasoning length appears to be the key factor,  a natural question arises: \textbf{How does increasing reasoning length impact performance at scale?} To explore this, the next section establishes a scaling law to quantify this relationship.

\section{The impact of reasoning length}
\subsection{The scaling law on reasoning length}
\subsubsection{Experimental Setup}

In the previous section, we showed that reasoning length has a greater impact on model performance than problem difficulty. The reasoning length of problems in the training dataset emerged as a key factor in performance improvement. To explore this further, we examined how variations in token length during reasoning affect model performance while keeping problem difficulty constant. Specifically, we grouped problems of the same difficulty into four levels based on reasoning token lengths: 1.5k, 3k, 6k, and 12k. For consistency, all training datasets were limited to 300 samples, and experiments were conducted using the Qwen2.5-32B-Instruct model. To illustrate the relationship between reasoning length and performance, we selected two representative reasoning benchmarks: MATH500, a dataset of 500 high-difficulty mathematical competition problems. GPQA Diamond, a multi-disciplinary benchmark designed to assess deep reasoning and domain-specific expertise.  These benchmarks were chosen to evaluate model performance across different reasoning lengths and complexities, providing strong evidence for the impact of reasoning length on performance.

\subsubsection{Result}

\begin{figure}[h]
	\centering
	\includegraphics[width=1\textwidth]{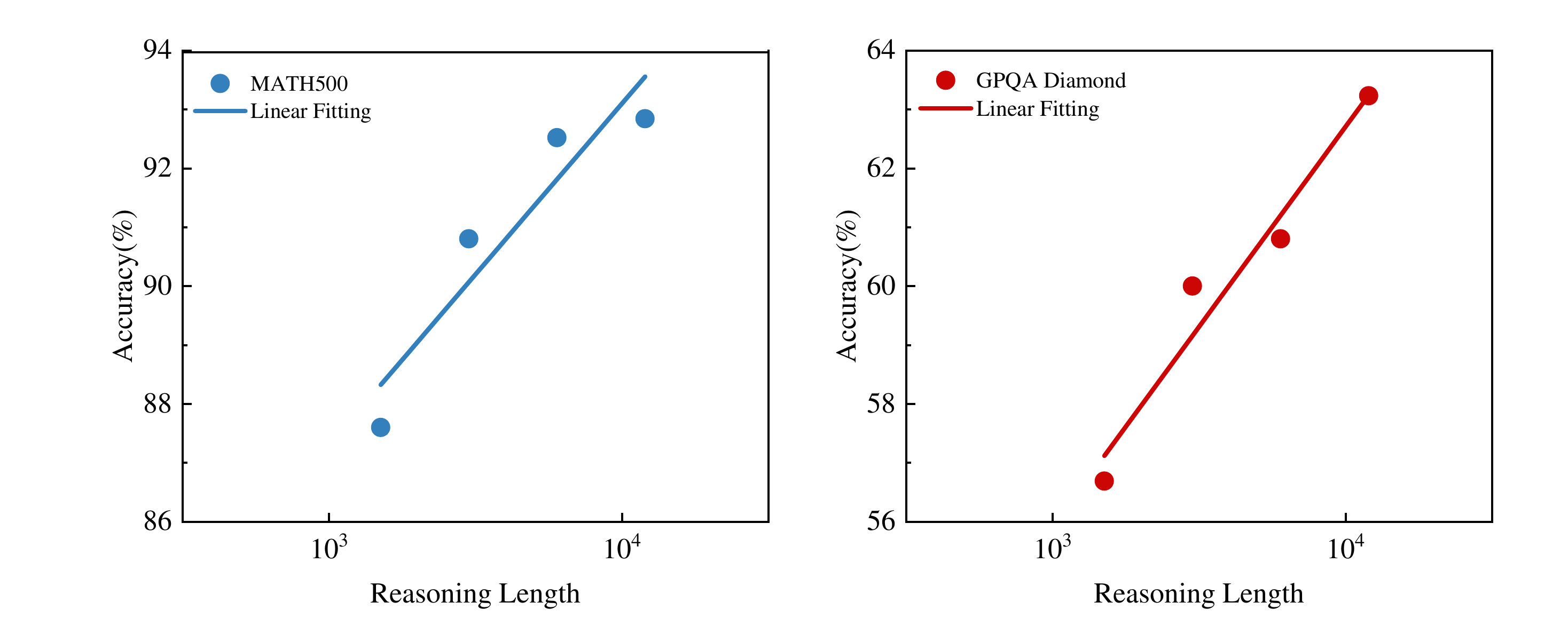}
	\caption{Test results of models trained under different reasoning length levels on MATH500 and GPQA Diamond.}
	\label{fig:scaling_law}
\end{figure}

Figure~\ref{fig:scaling_law}(a) shows the relationship between reasoning length and model accuracy on the MATH500 benchmark. The x-axis represents different reasoning length levels, while the y-axis indicates model accuracy as a percentage. The trend in the figure suggests that accuracy steadily increases from 87.6\% to nearly 93\% as reasoning length grows, following a clear linear pattern. Similarly, Figure~\ref{fig:scaling_law}(b) illustrates this relationship on the GPQA Diamond benchmark, where accuracy improves from 56.9\% to over 63\% with longer reasoning sequences.  

These results indicate that across both mathematical reasoning and other domains, as the reasoning length in training data increases exponentially, model performance improves in a nearly linear fashion. This highlights the efficiency gains achievable by extending reasoning sequences. The consistent improvement across both benchmarks further confirms that reasoning length plays a fundamental role in enhancing model performance. This finding suggests that reasoning length is a key factor influencing performance across diverse reasoning tasks, providing valuable insights for optimizing training strategies and dataset design.

\subsection{The reason for performance improvement}
To investigate the impact of reasoning length on model performance, we conducted experiments comparing two models trained with reasoning lengths of 1.5k and 12k. Using the MATH500 test set, we analyzed their reasoning processes, examining both successful and failed reasoning attempts. Our analysis included statistical comparisons of average reasoning token length and the top 10 most frequently used words during reasoning. The goal was to understand why the model trained with a reasoning length of 12k achieved over a 5\% improvement in accuracy.

As shown in Table~\ref{tab:2}, certain transition words such as "but" and "wait" frequently appear during reasoning, indicating the model's ability to reflect on and adjust its approach~\cite{Li2025FromS1}~\cite{Ye2025LIMOLI}. In successful reasoning cases, the frequency of these words remains around 2\% for both models. However, in failed cases, the model trained with a reasoning length of 1.5k shows a sharp increase in transition word usage to nearly 9\%, whereas the 12k model maintains a stable 2\%. This suggests that even when errors occur, the reasoning process of the 12k model remains more structured. This finding indicates that training with longer reasoning sequences helps the model internalize a more consistent reasoning framework, ultimately leading to improved accuracy.

Furthermore, when the reasoning length in training data increases by a factor of 8, the average token length of the 12k model’s reasoning process is roughly twice that of the 1.5k model, regardless of success or failure. This suggests that extending reasoning length does not exponentially increase computational cost, making it a practical approach for enhancing model performance.

\begin{table*}[h]
	\centering
	\setlength{\tabcolsep}{6pt}
	{
		\fontsize{10}{13}\selectfont
		\renewcommand\arraystretch{1.5}
		\resizebox{\linewidth}{!}{
			\begin{tabular}[l]{p{0.2\linewidth}p{0.15\linewidth}p{0.15\linewidth}p{0.7\linewidth}}
				\toprule
				\textbf{Dataset Size} & \textbf{Correct/ Wrong} & \textbf{Average Tokens} & \textbf{Top 10 Frequently Occurring Words}  \\ \midrule
				1.5k & Correct & 2147.65 & the(5.30\%) is(3.24\%) so(1.98\%) of(1.45\%) to(1.44\%) and(1.25\%) that(1.17\%) let(1.08\%) \textbf{wait(1.07\%)} \textbf{but(0.91\%)}\\ \midrule		
				12k & Correct & 4716.27 & the(4.92\%) is(3.04\%) so:(1.83\%) to:(1.41\%) of:(1.25\%) and:(1.19\%) \textbf{but:(1.19\%)} let:(0.93\%) that:(0.90\%) \textbf{wait:(0.81\%)}\\ \midrule
				1.5k & Wrong & 8247.21 & \textbf{but(5.05\%)} the(5.00\%) \textbf{wait(3.78\%)} is(3.24\%) of(1.29\%) so(1.26\%) therefore(1.16\%) to(1.08\%) and(1.01\%) that(0.70\%)\\ \midrule
				12k & Wrong & 15694.54 & the(5.12\%) is(2.85\%) to(1.64\%) and(1.42\%) \textbf{but(1.27\%)} of(1.20\%) so(1.08\%) \textbf{wait(0.80\%)} that(0.80\%) in(0.75\%)
				\\ \bottomrule
				\end{tabular}
			}
			\caption{Statistical analysis of models trained on datasets with varying reasoning lengths. The table presents three distinct metrics: Correct/Wrong, indicating whether the reasoning in the test was successful or failed; the average reasoning length of the test results; and the top 10 most frequent words appearing in the test results.}
			\label{tab:2}
			}
		\end{table*}

\section{Synthetic Reasoning Data Construction for Long1K}

\subsection{Data Source}

This section outlines the process of constructing the Long1K dataset for model training. The Long1K dataset is constructed following the approach used in Long Problem(Double). Long1k is composed of data from two main sources:
Openthouhts-114k~\cite{openthoughts} and s1k~\cite{Muennighoff2025s1ST}. The s1k dataset was created by collecting 59,029 problems from 16 sources, followed by a selection process based on quality, difficulty, and diversity. Ultimately, 1,000 high-quality and diverse problems were chosen, with their reasoning steps and answers generated using the Google Gemini Flash Thinking API. Meanwhile, Openthouhts-114k was compiled from nine sources, covering four categories: Code, Math, Science, and Puzzle. Reasoning steps for these problems were generated using DeepSeek-R1 and validated for accuracy to form the final 114k dataset.

\subsection{Synthetic Method}

Long1k contains 1000 pieces of data, among which 800 pieces of data come from Openthouhts114k. Specifically, on the one hand, we randomly selected two mathematical problems from Openthouhts114k. The questions, reasoning processes and results of these two maths problems were used to join together using different connectives to increase the length of the prompt. It is important to note that the length of the final synthesised prompt needs to float above and below the specified length.In addition, we restate and rewrite the connectives in terms of diversity to improve the accuracy and robustness of the model and reduce the dependence on prompts. Please refer to the following table~\ref{tab:5} for the specific prompt sample templates.

Specifically, to construct our dataset, we first randomly selected two mathematical problems from Openthouhts114k. The problem statements of these two mathematical problems served as the content for the User Prompt, which were concatenated using simple language. Their reasoning processes and results were used as the content for the System Prompt, encapsulated between the \textless\textbar begin\_of\_thought\textbar\textgreater and \textless\textbar end\_of\_thought\textbar\textgreater tags, as well as between the \textless\textbar begin\_of\_solution\textbar\textgreater and \textless\textbar end\_of\_solution\textbar\textgreater tags, and then connected using simple language, similar to “This is a conversation between a user and an assistant, where the user poses a question, and the assistant provides a reasoning process followed by an answer.” These parts were then connected using various conjunctions to increase the length of the prompt. It is important to note that the length of the final synthesised prompt needs to float above and below the specified length.In addition, we restate and rewrite the connectives in terms of diversity to improve the accuracy and robustness of the model and reduce the dependence on prompts. Please refer to the Appendix B for the specific prompt sample templates.

In order to avoid the overfitting problem of the model for this spliced type of questions, we also randomly extracted 200 maths questions from the s1.1 dataset and added them to the above spliced data. Similarly, we provided specific questions in the User Prompt section. In the System Prompt section, the reasoning process and the answer were enclosed within  \textless\textbar begin\_of\_thought\textbar\textgreater and \textless\textbar end\_of\_thought\textbar\textgreater tags, as well as \textless\textbar begin\_of\_solution\textbar\textgreater and \textless\textbar end\_of\_solution\textbar\textgreater tags, respectively. The sample prompt template is shown in Appendix B. Ultimately, the synthetic data Long1K used for model training will consist of these two parts of data.

\section{Experiment}

\subsection{Setup}
The experiments in this paper were conducted using 8 NVIDIA A800-SXM4-80GB. In some experiments 4 gpu cards are used. The number of epoch is set to 3, and the batch size is 2 on each gpu. In addition, the learning rate is set to 2e-4, and the cosine function is selected for the Learning Rate Scheduler. The LORA full parameter fine-tuning strategy is used, where the parameters lora\_rank, lora\_alpha are 16, 32 respectively. All of our experimental results are averages of five trials, including the results mentioned previously.

We fine-tuned Qwen2.5-32B-Instruct on the Long1K dataset to obtain our model Long1K-32B, with a maximum output length of 32k tokens. To evaluate its performance, three well-established reasoning benchmarks were selected: 

\begin{itemize}
	\item \textbf{AIME2024/2025:} All contain 30 mathematical problems, which are derived from a highly regarded high school mathematics competition in the United States, known for its complex problems that require strong problem-solving skills and deep mathematical reasoning. 
	\item \textbf{MATH500:} A collection of 500 mathematical problems spanning various difficulty levels and problem types, commonly used in research on mathematical problem-solving and natural language processing. 
	\item \textbf{GPQA Diamond:} A dataset consisting of 198 doctoral-level scientific questions across biology, chemistry, and physics.
\end{itemize}

Long1K-32B was compared against leading models across these benchmarks, focusing on two categories: highperforming models trained on datasets of similar size to ours, and exceptionally prominent models trained on significantly larger datasets.

\subsection{Main Result}
Our experimental results, presented in Table~\ref{tab:3}, demonstrate the exceptional capabilities of Long1K-32B.

\begin{itemize}
	\item \textbf{Comparable-Size Models:} Compared to other models trained on datasets of similar size, this model demonstrates strong performance across all benchmarks. It achieves an accuracy of 95.6\% on the MATH500 benchmark and 71.1\% on the GPQA Diamond benchmark, both of which are the highest among the models in the comparison.
	\item \textbf{Large-Scale Models:} When compared to models trained on significantly larger datasets, this model maintains strong performance, achieving results(71.1\%) on par with DeepSeek-R1(71.5\%) on the GPQA Diamond benchmark.
\end{itemize}

\begin{table*}[!t]
	\centering
	\setlength{\tabcolsep}{6pt}
	{
		\fontsize{10}{13}\selectfont
		\renewcommand\arraystretch{1.5}
		\resizebox{\linewidth}{!}{
			\begin{tabular}[l]{p{0.3\linewidth}p{0.15\linewidth}p{0.15\linewidth}p{0.15\linewidth}p{0.15\linewidth}p{0.2\linewidth}}
				\toprule
				\textbf{Model Name} & \textbf{Dataset Size} & \textbf{MATH500} & \textbf{AIME2024} & \textbf{AIME2025} & \textbf{GPQA Diamond}   \\ \midrule
				\textbf{s1-32B} & 1k & 92.6 & 50.0 & 26.7 & 56.6 \\ 		
				\textbf{s1.1-32B} & 1k & 89.0 & 64.7 & 49.3 & 60.1 \\ 
				\textbf{LIMO} & 0.8k & \underline{94.8} & 57.1 & 49.3 & \underline{66.7} \\ \midrule
				\textbf{OpenThinker-32B} & 114k & 90.6 & \underline{66.0} & \underline{53.3} & 61.6 \\
				\textbf{DeepSeek-R1-Distill-Qwen-32B} & 800k & 93.0 & \textbf{72.6} & \textbf{55.9} & 62.1 \\ \midrule
				\textbf{Long1K-32B} & 1k & \textbf{95.6} & 50.7 & \underline{53.3} & \textbf{71.1} \\ \bottomrule
			\end{tabular}
		}
		\caption{Performance comparison of different models across multiple reasoning benchmarks (pass@1). The best results for each benchmark are highlighted in bold, with the second-best underlined. The data for s1 does not use budget forcing, and the data for s1.1 that does not use budget forcing comes from Open Thoughts~\cite{openthoughts}. Among models trained on datasets of similar scale, Long1K-32B performs exceptionally well. It maintains strong competitiveness even against models trained on substantially larger datasets~\cite{Zeng2025SIFTGL}, showing particularly notable results on the MATH500 and GPQA Diamond benchmarks.}
		\label{tab:3}
	}
\end{table*}

Experimental results show that the approach of splicing multiple problem descriptions and their solution processes through connectives can effectively extend the inference length and thus significantly improve the inference performance of the model. Compared to increasing problem difficulty, which often requires complex task design and data filtering, this approach offers a easier and more straightforward strategy at the data level. It also suggests that optimising inference length, rather than just increasing problem difficulty, can be a more effective way of enhancing model capability. This provides a more flexible approach for future training and dataset construction.

\section{Related works}
\subsection{Long-thought reasoning}
Studies have shown that as model size increases, LLMs exhibit significant improvements in long-thought reasoning tasks.~\cite{Brown2020LanguageMA}~\cite{Wei2022EmergentAO}. In the field of long-thought reasoning, researchers have proposed a variety of powerful models, such as ChatGPT O1~\cite{jaech2024openai}, DeepSeek-R1~\cite{guo2025deepseek} and QwQ-32B~\cite{qwq32b} which have demonstrated remarkable performance in handling complex reasoning tasks. The success of these models is partly attributed to their large number of parameters and extensive training data, enabling them to capture complex reasoning patterns~\cite{Hoffmann2022TrainingCL}. These models typically simulate human cognitive processes through advanced techniques in deep learning and natural language processing. In terms of model reproduction, two primary approaches have emerged: knowledge distillation and reinforcement learning from human feedback (RLHF)~\cite{deepscaler2025}.The former compresses the knowledge of large models into smaller ones, significantly reducing computational costs while maintaining high reasoning performance~\cite{Hinton2015DistillingTK}. As exemplified by SkyThought~\cite{Li2025LLMsCE}, elevates LLM reasoning efficiently with Long CoT data. Our study proposes a novel approach based on the distilled DeepSeek-R1~\cite{guo2025deepseek}, aiming to improve reasoning capability. Although currently grounded in distilled models, the proposed framework exhibits potential for extension to RLHF-based models, thereby offering new directions for the future development of long-thought reasoning systems.
\subsection{Challenges in Long-thought Reasoning: Difficulty vs. Length}
In existing research on mathematical reasoning~\cite{Ballon2025TheRB}~\cite{Shao2024DeepSeekMathPT}, difficulty has been widely regarded as a critical factor for training models, with studies such as s1~\cite{Muennighoff2025s1ST} collects questions following three guiding principles: quality, difficulty and diversity. However, our work challenges this paradigm by demonstrating that length, rather than difficulty, is a more significant factor in long-thought reasoning tasks. Unlike prior approaches that rely on artificially creating challenging problems~\cite{Ding2024UnleashingRC}~\cite{Luo2023WizardMathEM}, our research shifts the focus from the difficulty of problems to the length of reasoning chains. We propose a simple and effective method of concatenating existing problems to extend reasoning chains. This approach not only reduces the complexity of problem design but also aligns better with real-world scenarios where reasoning often involves extended sequences of thought. 
\subsection{Scaling law on LLM}
Scaling laws have emerged as a crucial tool in the research of large language models (LLMs) for guiding model design and predicting performance. During the training phase, researchers~\cite{Kaplan2020ScalingLF} have identified that model performance scales predictably with model size, dataset size, and computational resources. In the inference phase, the focus of Scaling Law research has shifted towards optimizing inference compute. Studies~\cite{Bi2024DeepSeekLS}~\cite{Snell2024ScalingLT}~\cite{Bi2024DeepSeekLS} show that increasing inference compute can enhance inference performance without significantly increasing model parameters , and it is proposed that optimizing inference strategies can significantly improve inference accuracy. Our research introduces data length as a crucial scaling factor, demonstrating that the performance of LLMs in multi-step mathematical reasoning scales significantly with the length of reasoning chains.

\section{Conclusion}

This study addresses the key question of which of problem difficulty and reasoning length has a greater impact on the reasoning performance of LLM. Existing studies have generally hypothesised that difficult tasks enable LLM to make more effective use of their pre-trained knowledge by inducing more complex inference chains~\cite{guo2025deepseek}. This paper proposes that inference length, rather than task difficulty, may be the key factor in improving model performance. While difficult tasks do tend to generate longer inference chains, our study suggests that inference length plays a more critical role in model performance improvement.

Through a synthetic approach, we generate Long1K datasets with longer inference lengths. Based on this, we fine-tuned to obtain the Long1K-32B model and achieved state-of-the-art performance on benchmarks such as MATH500 and GPQA Diamond. The results show that an increase in inference length significantly improves the model's performance on complex problems, while an increase in task difficulty is not always directly related to performance improvement. This finding challenges the conventional view and provides new perspectives for understanding the inference mechanism of LLM. In addition, we explore the relationship between inference length and the efficiency of model knowledge utilisation, and find that longer inference chains may help models to better integrate and invoke their pre-trained knowledge. At the same time, this also effectively improves the model's reasoning ability in dealing with mathematical problems, thus showing stronger generalisation ability in complex tasks.

This work not only provides a theoretical basis for optimising the task design of LLM, but also points out the direction for future research. For example, how to further improve the model performance by controlling the inference length, and how to balance the relationship between inference length and task difficulty in different task types are issues that deserve in-depth investigation. In addition, our study suggests that future model training and evaluation methods may need to focus more on the optimisation of the reasoning process rather than just the difficulty design of the task itself.

\bibliographystyle{elsarticle-harv}
\bibliography{lbs}

\newpage
\appendix

\section{Extracting mathematical knowledge points and synthesizing questions}

We use the following prompt to extract knowledge points from mathematical problems. These knowledge points are then used to combine long and difficult questions, with the prompts used for synthesis also shown here. To ensure fairness in the experiment, each question is synthesized using 4 knowledge points.

\begin{figure}[h!]
	\centering
	\includegraphics[width=1\textwidth]{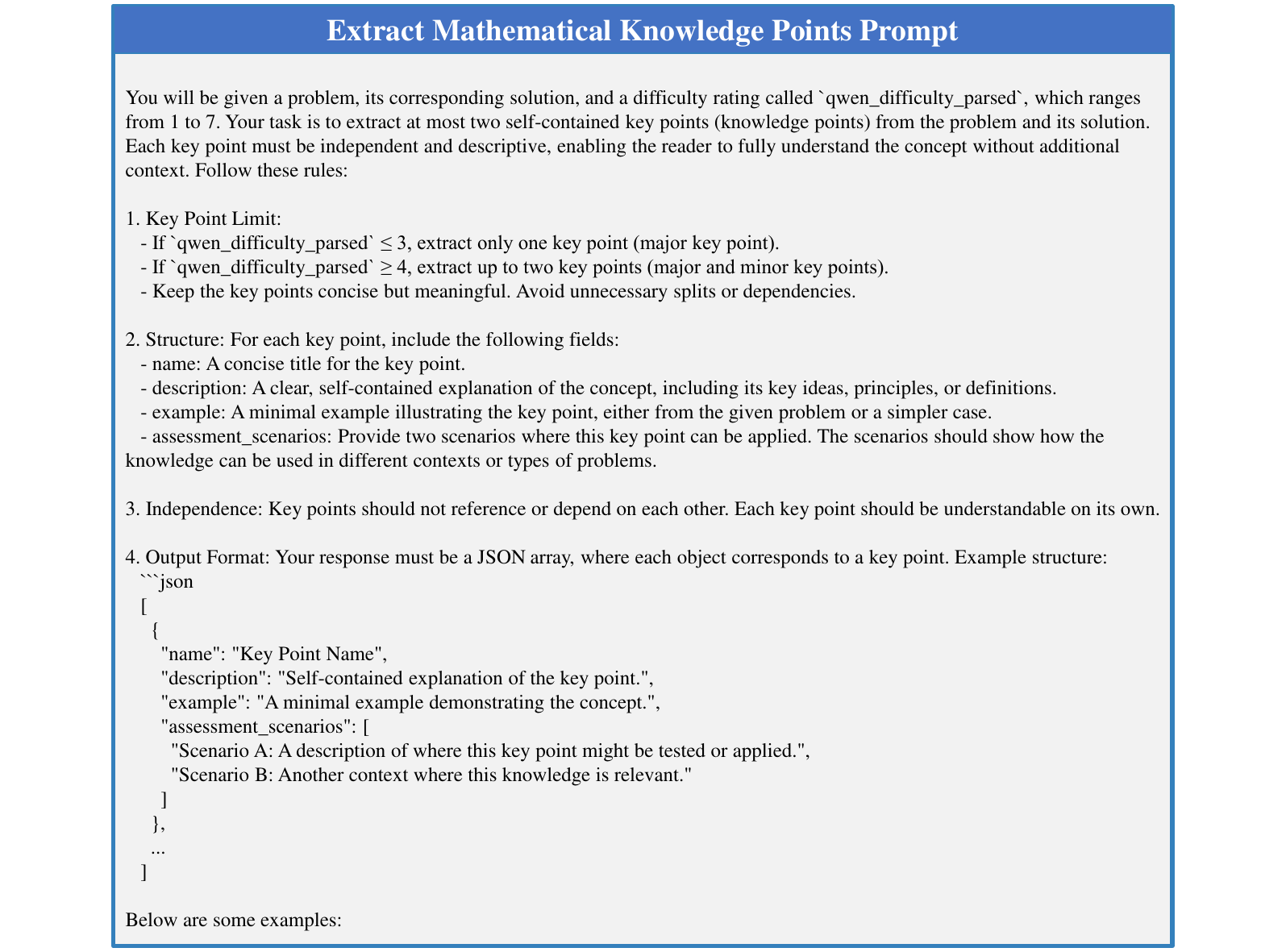}
	\caption{This is the prompt for extracting mathematical knowledge points.}
	\label{fig:extract_prompt}
\end{figure}

\newpage

\begin{figure}[h!]
	\centering
	\includegraphics[width=1\textwidth]{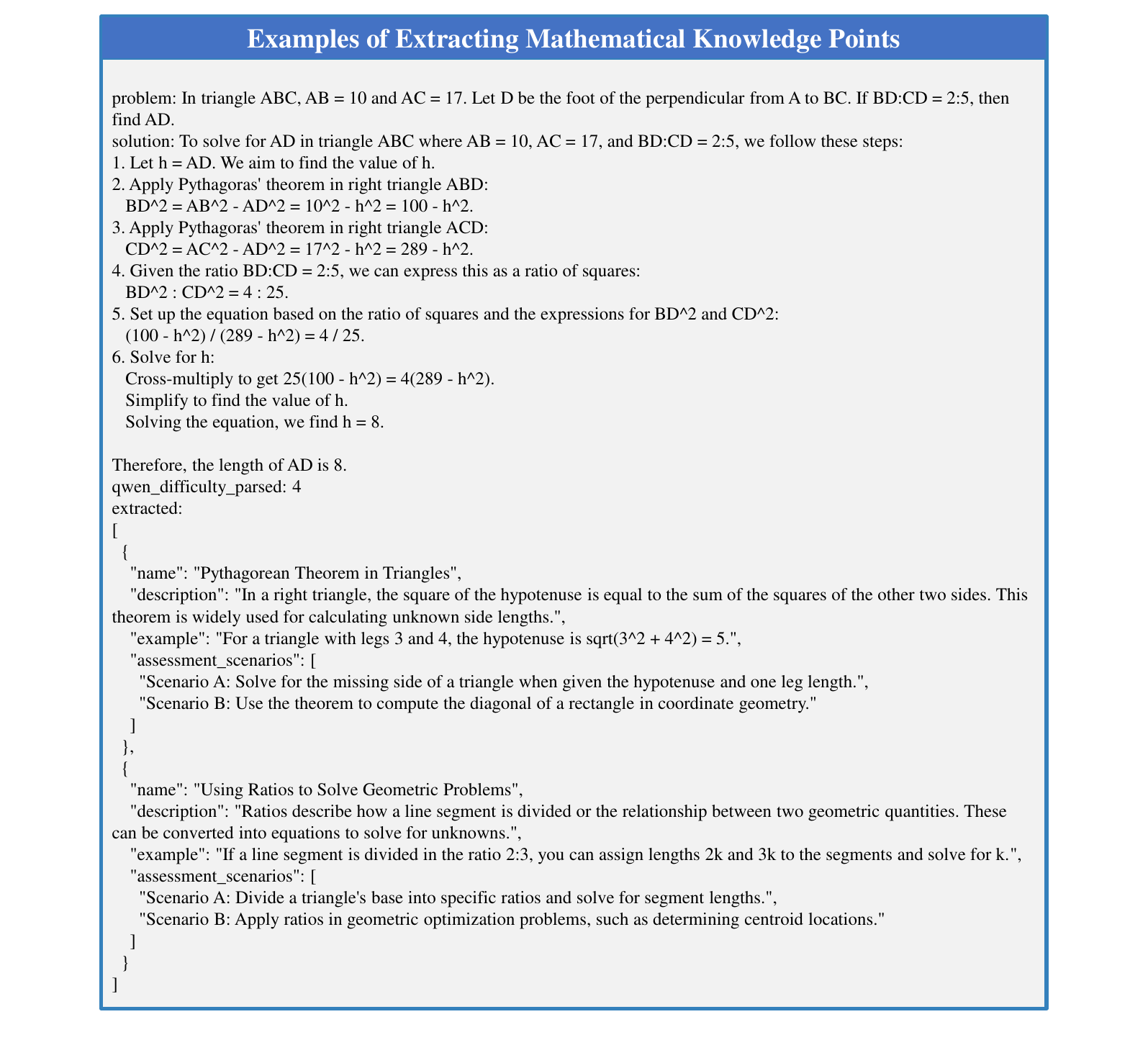}
	\caption{This is an example of extracting mathematical knowledge points.}
	\label{fig:example}
\end{figure}

\newpage

\begin{figure}[h!]
	\centering
	\includegraphics[width=1\textwidth]{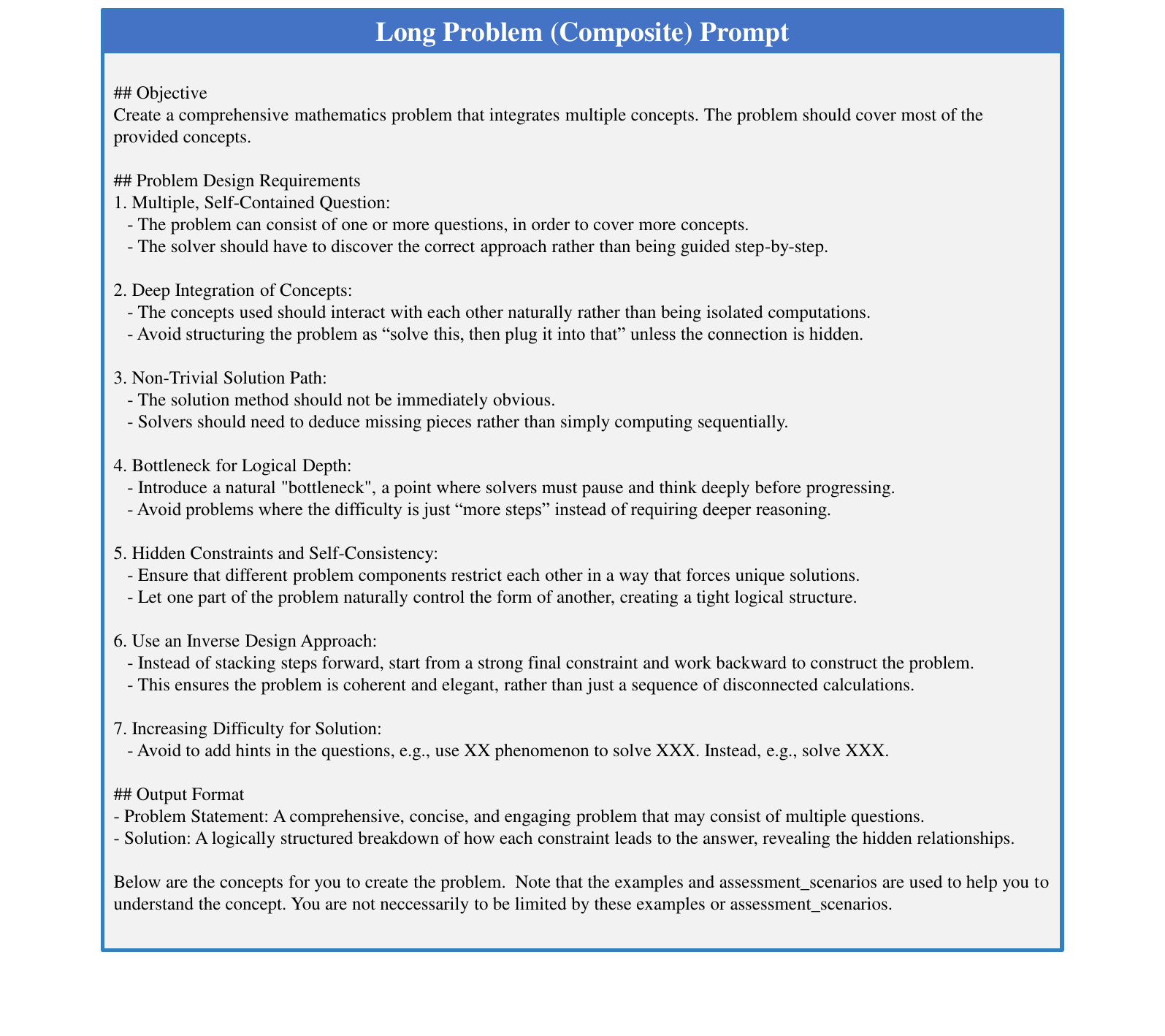}
	\caption{This is a prompt for Long Problem (Composite) based on multiple mathematical concepts.}
	\label{fig:long_problems_prompt}
\end{figure}

\newpage

\begin{figure}[h!]
	\centering
	\includegraphics[width=1\textwidth]{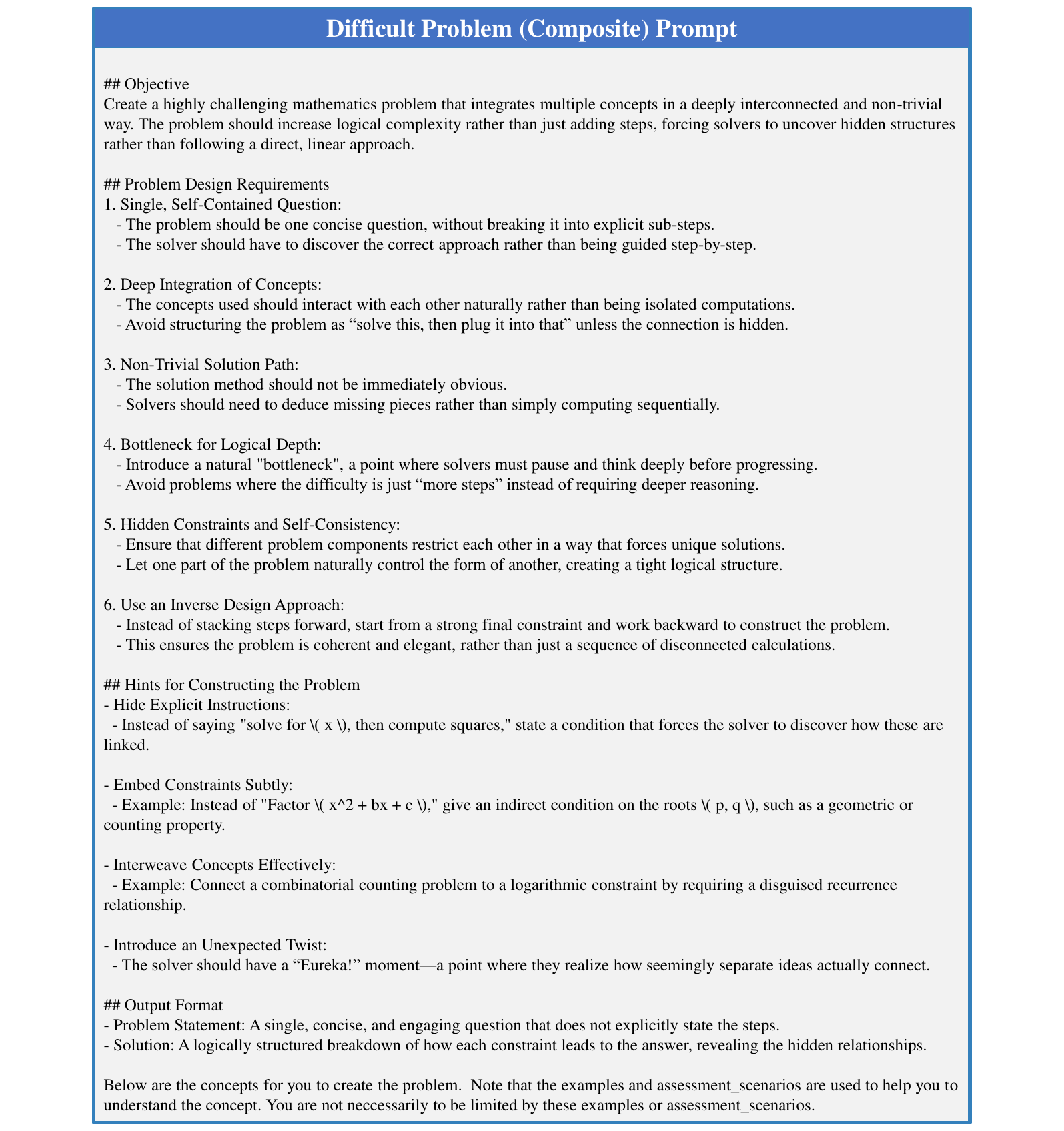}
	\caption{This is a prompt for Difficult Problem (Composite) based on multiple mathematical concepts..}
	\label{fig:difficultt_problem_prompt}
\end{figure}

\newpage
\section{Prompt for synthesizing Long1k.}

\begin{table*}[h]
	\centering
	\setlength{\tabcolsep}{6pt}
	{
		\fontsize{10}{13}\selectfont
		\renewcommand\arraystretch{1.5}
		\resizebox{\linewidth}{!}{
			\begin{tabular}{p{0.2\linewidth}| p{0.9\linewidth}}
				\toprule
				\textbf{System Prompt} & \textless\textbar im\_start\textbar\textgreater system$\backslash$n
				
				A conversation between User and Assistant. The user asks a question, and the Assistant solves it. The assistant first thinks about the reasoning process in the mind and then provides the userwith the answer. The reasoning process and answer are enclosed within \textless\textbar begin\_of\_thought\textbar\textgreater \textless\textbar end\_of\_thought\textbar\textgreater and \textless\textbar begin\_of\_solution\textbar\textgreater \textless\textbar end\_of\_solution\textbar\textgreater tags, respectively, i.e., \textless\textbar begin\_of\_thought\textbar\textgreater reasoning process here \textless\textbar end\_of\_thought\textbar\textgreater \textless\textbar begin\_of\_solution\textbar\textgreater answer here \textless\textbar end\_of\_solution\textbar\textgreater.
				
				\textless\textbar im\_end\textbar\textgreater$\backslash$n
				\\ \midrule
				\textbf{User Prompt} & \textless\textbar im\_start\textbar\textgreater user$\backslash$n
				
				I need help with two problems.The first one is \textbf{PROBLEM1}, and the second one is \textbf{PROBLEM2}.
				
				\textless\textbar im\_end\textbar\textgreater$\backslash$n
				\\ \midrule		
				\textbf{Output} & \textless\textbar im\_start\textbar\textgreater system$\backslash$n
				
				\textless\textbar begin\_of\_thought\textbar\textgreater$\backslash$n$\backslash$n I will handle these two problems one by one. $\backslash$nI will start with the first problem. \textbf{THINGKING1}.$\backslash$n Now I will turn to the second problem. \textbf{THINGKING2}. \textless\textbar end\_of\_thought\textbar\textgreater$\backslash$n$\backslash$n
				\textless\textbar begin\_of\_solution\textbar\textgreater$\backslash$n$\backslash$n  The solution for the first problem is as belows. \textbf{SOLUTION1}.$\backslash$n The solution for the second problem is as belows. \textbf{SOLUTION2}. \textless\textbar end\_of\_solution\textbar\textgreater
				
				\textless\textbar im\_end\textbar\textgreater
				\\ \bottomrule
			\end{tabular}
		}
		\caption{Sample prompt template for splicing maths problems in Long1k. The red parts of it need to be replaced with the appropriate content.}
		\label{tab:5}
	}
\end{table*}

\begin{table*}[!b]
	\centering
	\setlength{\tabcolsep}{6pt}
	{
		\fontsize{10}{13}\selectfont
		\renewcommand\arraystretch{1.5}
		\resizebox{\linewidth}{!}{
			\begin{tabular}{p{0.2\linewidth}| p{0.9\linewidth}}
				\toprule
				\textbf{System Prompt} & \textless\textbar im\_start\textbar\textgreater system$\backslash$n
				
				A conversation between User and Assistant. The user asks a question, and the Assistant solves it.The assistant first thinks about the reasoning process in the mind and then provides the userwith the answer. The reasoning process and answer are enclosed within \textless\textbar begin\_of\_thought\textbar\textgreater \textless\textbar end\_of\_thought\textbar\textgreater and \textless\textbar begin\_of\_solution\textbar\textgreater \textless\textbar end\_of\_solution\textbar\textgreater tags, respectively, i.e., \textless\textbar begin\_of\_thought\textbar\textgreater reasoning process here \textless\textbar end\_of\_thought\textbar\textgreater \textless\textbar begin\_of\_solution\textbar\textgreater answer here \textless\textbar end\_of\_solution\textbar\textgreater.
				
				\textless\textbar im\_end\textbar\textgreater$\backslash$n
				\\ \midrule
				\textbf{User Prompt} & \textless\textbar im\_start\textbar\textgreater user$\backslash$n
				
				Help me to solve the following problem. \textbackslash" \textbf{PROBLEM} \textbackslash".
				
				\textless\textbar im\_end\textbar\textgreater$\backslash$n
				\\ \midrule		
				\textbf{Output} & \textless\textbar im\_start\textbar\textgreater system$\backslash$n
				
				\textless\textbar begin\_of\_thought\textbar\textgreater$\backslash$n$\backslash$n The thinking process for problem as follows. \textbf{THINGKING}   
				
				\textless\textbar end\_of\_thought\textbar\textgreater$\backslash$n$\backslash$n
				
				\textless\textbar begin\_of\_solution\textbar\textgreater$\backslash$n$\backslash$n  \textbf{SOLUTION} \textless\textbar end\_of\_solution\textbar\textgreater 
				
				\textless\textbar im\_end\textbar\textgreater
				\\ \bottomrule
			\end{tabular}
		}
		\caption{Sample prompt template for a single maths problem in Long1k.}
		\label{tab:6}
	}
\end{table*}

\end{document}